\documentclass[sigconf]{acmart}

\renewcommand\footnotetextcopyrightpermission[1]{} 
\usepackage{fancyhdr}
\usepackage{color,xcolor}
\usepackage{array}
\usepackage{multirow}
\usepackage{algorithm}
\usepackage{algorithmic}
\usepackage{epsfig}
\usepackage{hyperref}
\usepackage{booktabs}
\usepackage{threeparttable}

\newcommand{\ignore}[1]{}

\begin{document}

\title{Invisible Backdoor Attacks Using Data Poisoning in the Frequency Domain}


\author{Chang Yue}
\affiliation{%
  \institution{SKLOIS, Institute of Information Engineering, Chinese Academy of Sciences}
  \city{Beijing}
  \country{China}}
\email{yuechang@iie.ac.cn}

\author{Peizhuo Lv}
\affiliation{%
  \institution{SKLOIS, Institute of Information Engineering, Chinese Academy of Sciences}
  \city{Beijing}
  \country{China}}
\email{lvpeizhuo@iie.ac.cn}

\author{Ruigang Liang}
\affiliation{%
  \institution{SKLOIS, Institute of Information Engineering, Chinese Academy of Sciences}
  \city{Beijing}
  \country{China}}
\email{liangruigang@iie.ac.cn}

\author{Kai Chen}
\affiliation{%
  \institution{SKLOIS, Institute of Information Engineering, Chinese Academy of Sciences}
  \city{Beijing}
  \country{China}}
\email{chenkai@iie.ac.cn}





\begin{abstract}
With the rapid development and broad application of deep neural networks (DNNs), backdoor attacks have gradually attracted attention due to their great harmfulness. Backdoor attacks are insidious, and poisoned models perform well on benign samples and are only triggered when given specific inputs, which then cause the neural network to produce incorrect outputs. The state-of-the-art backdoor attack work is implemented by data poisoning, i.e., the attacker injects poisoned samples (some samples patched with a trigger) into the dataset, and the models trained with that dataset are infected with the backdoor. However, most of the triggers used in the current study are fixed patterns patched on a small fraction of an image and are often clearly mislabeled, which is easily detected by humans or some defense methods such as Neural Cleanse and SentiNet. Also, it is difficult to be learned by DNNs without mislabeling, as they may ignore small patterns.

In this paper, we propose a generalized backdoor attack method based on the frequency domain, which can implement backdoor implantation without mislabeling the poisoned samples and accessing the training process. It is invisible to human beings and able to evade the commonly used defense methods. 
We evaluate our approach in the no-label and clean-label cases on three benchmark datasets (CIFAR-10, STL-10, and GTSRB) with two popular scenarios, including self-supervised learning and supervised learning. The results show that our approach can achieve a high attack success rate (above 90\%) on all the tasks without significant performance degradation on main tasks.
Also, we evaluate the bypass performance of our approach for different kinds of defenses, including the detection of training data (i.e., Activation Clustering), the preprocessing of inputs (i.e., Filtering), the detection of inputs (i.e., SentiNet), and the detection of models (i.e., Neural Cleanse). The experimental results demonstrate that our approach shows excellent robustness to such defenses. 


\end{abstract}




\maketitle

\section{Introduction}

With the significant improvement of computing power, deep learning has been developed rapidly,
in particular, supervised learning trained on labeled datasets and self-supervised learning trained on pretext tasks with unlabeled datasets, which have been widely applied in various areas and have profoundly changed people's production and lifestyle, such as face recognition~\cite{Schroff2015FaceNetAU,parkhi2015deep,Wiles2018SelfsupervisedLO}, speech recognition~\cite{BAI202165,Tao2021SelfsupervisedSR}, autonomous vehicles~\cite{ap,Luo2021SelfSupervisedPM}, and remote diagnosis~\cite{2018FDA}. 

The ubiquitous and successful application of deep learning in various fields simultaneously brings new security issues, such as adversarial attacks~\cite{2013Intriguing,Xiao2018GeneratingAE,zhao2019seeing} and backdoor attacks~\cite{Gu2019BadNetsEB,Liu2018TrojaningAO,Chen2017TargetedBA}. Unlike adversarial attacks that explore the intrinsic vulnerability of DNNs in the inference phase, backdoor attacks always poison models in the training phase, which could perform well on benign samples but output false output when fed backdoor samples. 
For example, suppose a company's face recognition access control system suffers a backdoor attack (e.g., a pair of glasses with a unique shape can trigger the backdoor). In that case, the system will recognize the adversary wearing the glasses as an employee within the company, posing a potentially serious security risk.



\begin{figure}[!t]
\centering
\epsfig{figure=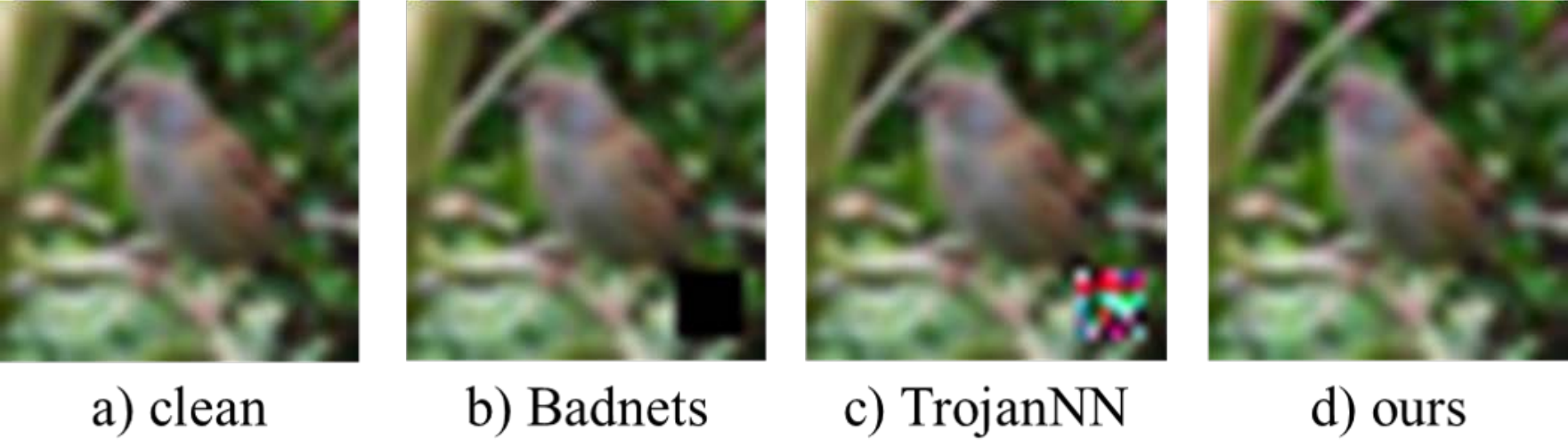, width=0.45\textwidth} 
\caption{The examples of backdoor attacks. a) is the original image, b), c), and d) are images patched with the trigger proposed by Badnets~\cite{Gu2019BadNetsEB}, TojanNN~\cite{Liu2018TrojaningAO} and our work. Particularly, Badnets and TrojanNN samples are always labeled to the target label (e.g., car), but ours is with a clean label.}
\label{fig:trigger}
\vspace{-7pt}
\end{figure}

State-of-the-art data poisoning-based work faces the bottleneck of insufficient stealthiness or poor robustness, as shown in Figure~\ref{fig:trigger}. 
1) the poisoned samples have fixed trigger patterns and wrong labels in labeled datasets and fixed trigger patterns in the unlabelled dataset, which can be effortlessly perceived by human beings. 2) the fixed trigger can be detected and reconstructed by some defenses, such as Neural Cleanse~\cite{Wang2019NeuralCI}, SentiNet~\cite{Chou2020SentiNetDL}.
Inspired by~\cite{Xu2019TrainingBO, Xu2020FrequencyPF, Luo2021TheoryOT}, i.e., DNN models can learn the signals in the frequency domain and a slight change in the frequency domain can influence all the pixels in the spatial domain that are invisible to humans, we believe that the frequency domain-based backdoor attack approach can solve the problems of the previous study mentioned above. However, there are some challenges in designing the frequency domain-based backdoor attack as below.

\vspace{3pt}\noindent\textbf{Challenges in designing frequency domain backdoor.}
C1: Be robust to defenses that preprocess the input, e.g., filters. 
C2: Bypass defenses based on trigger detection. The state-of-the-art backdoor defenses are based on detecting a specific trigger pattern in the spatial domain. C3: Simultaneously ensure the learnability and stealthiness of the frequency backdoor.
Trigger with higher frequencies and intensities is more accessible for DNN models to learn but more able to be perceived by humans.

In this paper, we propose an adaptive trigger selection algorithm to deal with the challenges above, which contains three phases.
In the first phase, we select multiple frequencies robust to some commonly used filters (e.g., Gaussian Filter) as candidates  (address C1).
In the second phase, we select frequencies that could make the trigger have different patterns on different images in the spatial domain to apply our modification in the frequency domain. Further, the more significant the difference between the trigger patterns among all the images, the more likely our attack will bypass such defenses.  (address C2).
In the third phase, we choose a target intensity whose value is slightly more significant than the average intensity among the original images but no greater than a threshold at each frequency selected, considering the frequency's location and the value of the average intensity (address C3). 

We evaluate our attack in two neural network learning scenarios, including self-supervised and supervised learning. 
For self-supervised learning, we pre-train ResNet-18~\cite{He2016DeepRL} on the poisoned CIFAR-10~\cite{krizhevsky2009learning} dataset (i.e., some images are patched with our frequency trigger) as the feature extractor using the popular methods SimCLR~\cite{Chen2020ASF} and MOCO V2~\cite{Chen2020ImprovedBW}, and then transfer to the downstream tasks, including CIFAR-10, STL-10~\cite{Coates2011AnAO} and GTSRB~\cite{stallkamp2011german}. We train ResNet-18 on the poisoned CIFAR-10 and STL-10 datasets for supervised learning. The experimental result demonstrates that our attack achieves over 90\% success rate on the poisoned samples of all the datasets, and only about 2\% performance degradation on the main tasks is incurred. Furthermore, we use PSNR (Peak Signal-to-Noise Ratio) and SSIM (Structural Similarity) to evaluate the changes to the original images due to our frequency trigger. The average PSNR on CIFAR-10 and STL-10 datasets is 24.11 and 26.94, respectively, and the average SSIM is 0.9024 and 0.9044, proving that our trigger has good stealthiness. In addition, We evaluate the bypass effectiveness of our attack against common backdoor detection methods, and the experimental result shows that our approach can bypass them all with high robustness.

\vspace{3pt}\noindent\textbf{Contributions}. Our main contributions are outlined below:

\noindent$\bullet$\space We propose a new invisible backdoor attack by designing a frequency trigger based on the statistical characteristics in the frequency domain. 
To the best of our knowledge, we are the first to achieve backdoor implantation on both no-label (i.e., self-supervised learning) and clean-label (i.e., supervised learning) scenarios without mislabeling the poisoned samples and accessing the training process.

\noindent$\bullet$\space We design an adaptive algorithm to choose appropriate properties for our frequency trigger, proving to make the trigger stealthier and more robust to the current commonly used defense methods.

\noindent$\bullet$\space We successfully implement an invisible backdoor attack in the frequency domain with over 90\% attack success rate while guaranteeing the performance of the main task.
\section{Background and Related Work}
\subsection{Spatial Domain and Frequency Domain} \label{2.1}
The spatial domain, also called image space, is a domain in which the pixels of the original image can be manipulated directly. RGB and YUV are two common color spaces used to record or display color images in the spatial domain. Both of them are based on the perceptual capabilities of the human eyes. RGB describes the combination of red, green, and blue, while YUV represents luminance (denoted as ``Y'' channel) and chrominance (denoted as ``U'' and ``V'' channels) that describes the light intensity and carries color information. Since the YUV color space is closer to human vision than RGB and has better perceptual quality~\cite{Podpora2014YUVVR}, YUV is increasingly used in the field of image processing~\cite{Xu2020LearningIT}.

The frequency domain provides a new perspective for image processing.
In the frequency domain, the low-frequency components correspond to smooth regions in the image, and the high-frequency components correspond to edges in the image.
The spatial domain and frequency domain can be transformed into each other by Fourier transform. DFT (Discrete Fourier Transform) and DCT (Discrete Cosine Transform) are the most classical transform methods. DCT is developed from DFT and is widely used in digital image processing because it has better energy aggregation in the frequency domain than DFT. The two-dimensional DCT and Inverse Discrete Cosine Transform (IDCT) are shown below.



\begin{equation}
\label{DCT}
F(u,v)=\sum_{i=0}^{M-1}\sum_{j=0}^{N-1}f(i,j)G(i,j,u,v)
\end{equation} 

\begin{equation}
\label{IDCT}
f(i,j)=\sum_{u=0}^{M-1}\sum_{v=0}^{N-1}F(u,v)G(i,j,u,v)
\end{equation} 
where,
\begin{equation}
G(i,j,u,v)=c(u)c(v)\cos{[\frac{(i+0.5)\pi}{M}u]}\cos{[\frac{(j+0.5)\pi}{N}v]}
\end{equation} 
and
\begin{flalign}
    c(u)=\begin{cases}
    \sqrt{\frac{1}{M}},u=0\\
    \sqrt{\frac{2}{M}},u\ne0
         \end{cases}
    & \ \ \ c(v)=\begin{cases}
    \sqrt{\frac{1}{N}},v=0\\
    \sqrt{\frac{2}{N}},v\ne0
         \end{cases}
\end{flalign}
where DCT transforms an image of size $M \times N$ from the spatial domain to the frequency domain of the same size (equation~\ref{DCT}), and IDCT does the opposite (equation~\ref{IDCT}). $F(u,v)$ represents the intensity at $(u,v)$ in the frequency domain, and $f(i,j)$ represents the pixel values at $(i,j)$ in the color space. 


\subsection{Backdoor Attacks}
\vspace{2pt}\noindent\textbf{Backdoor attacks in supervised learning.} 
In the last decade, deep supervised learning has achieved great success and has been involved in various complicated tasks, including computer vision tasks \cite{Redmon2016YouOL,Ren2015FasterRT,Schroff2015FaceNetAU}, natural language process \cite{Gardner2017AllenNLP,Melamud2016context2vecLG,Mikolov2013DistributedRO}, graph learning \cite{Wang2019DynamicGC,Kipf2017SemiSupervisedCW}, and so on. 
Unfortunately, with the popularity of DNNs, they are suffering from many
attacks. Backdoor attacks are mainly achieved by data poisoning, i.e., attackers train the victim DNNs in a poisoned training dataset with poisoned samples stamped with a trigger pattern (e.g., a small patch in the right-bottom corner of an image) relabelled to the target class.
To enhance the stealthiness of the backdoor attack, on the one hand, several works have proposed invisible backdoor trigger generation approaches. Chen et al.~\cite{Chen2017TargetedBA} propose a blended injection strategy to make the pattern of the trigger blend with the background image better.
Barni et al.~\cite{Barni2019ANB} choose a sinusoidal signal (i.e., fringe patterns in the spatial domain) as the backdoor trigger. Liu et al.~\cite{Liu2020ReflectionBA} utilize the natural reflection phenomenon to act as trigger. 
On the other hand, some studies try to achieve a backdoor attack in the clean-label scenario, preventing detection due to mislabeling. Turner et al.~\cite{Turner2018CleanLabelBA} first implement and discuss successful backdoor attacks under the clean label by utilizing adversarial examples and GAN-generated data. Barni et al.~\cite{Barni2019ANB} add a global ramp signal on the uniform dark background of MNIST dataset~\cite{lecun1998gradient} to make it detectable by the DNNs. Both results demonstrate that triggers that can cause global perturbations to the original images will be needed to achieve clean-label backdoor attacks.

\vspace{2pt}\noindent\textbf{Backdoor attacks in self-supervised learning.}
Unlike supervised learning, self-supervised learning trains the encoder on the pretext of tasks that leverage input data as supervision to help the encoder learn some critical features of the dataset. CLIP~\cite{Radford2021LearningTV} is trained on a wide variety of images with a wide variety of natural language supervision, which are abundantly available on the internet. 
SimCLR~\cite{Chen2020ASF} is proposed to learn the visual representations by contrastive learning, which learns representations by maximizing the similarity between differently augmented representations of the same example via a contrastive loss in the latent space. Inspired by SimCLR, MoCo v2~\cite{Chen2020ImprovedBW} improves the Momentum Contrast self-supervised learning algorithm by introducing blur augmentation and outperforming the SimCLR with smaller batch sizes and fewer epochs.
At the same time,  backdoor attacks against self-supervised learning have begun to gain widespread attention. Carlini et al.~\cite{Carlini2021PoisoningAB} propose a backdoor attack against CLIP by patching a trigger on the images and modifying the corresponding text descriptions. Jia et al.~\cite{Jia2021BadEncoderBA} propose a backdoor attack on the pre-trained encoder by designing an optimization function that aggregates the feature vectors of images with embedded triggers into the output space of the encoder. Saha et al.~\cite{Saha2021BackdoorAO} utilize the training characteristics of contrastive learning to inject a backdoor to the model by patching a trigger on the images of the target class.


\vspace{2pt}\noindent\textbf{Backdoor attacks in frequency domain.}
Based on our observation, the frequency domain has natural stealthiness. Specifically, the slight changes in the frequency domain can cause invisible changes in the spatial domain, which are difficult to perceive by humans~\cite{Sonka1993ImagePA}.
Previous studies have demonstrated that DNN models can perceive information in the frequency domain and that it tends to learn components from low-frequency to high-frequency~\cite{Xu2019TrainingBO, Xu2020FrequencyPF, Luo2021TheoryOT}. Besides, Wang et al.~\cite{Wang2020HighFrequencyCH} and Yin et al.~\cite{Yin2019AFP} discuss the relation between the robustness, generalization of a DNN model and the frequency properties of training data in the frequency domain. Gueguen et al.~\cite{Gueguen2018FasterNN} and Xu et al.~\cite{Xu2020LearningIT} directly extract features from the frequency domain to classify images. 
Zeng et al.~\cite{Zeng2021RethinkingTB} analyze the frequency-domain characteristics of the current trigger, and they find that images patched with different triggers contain vital high-frequency components compared to the spectrum of clean images.

With the development of neural network interpretability in the frequency domain, several recent works have begun to study backdoor attacks from a frequency domain perspective. Wang et al.~\cite{Wang2021BackdoorAT} tries to inject the backdoor directly through the frequency domain. However, the proposed frequency trigger has a fixed pattern in the spatial domain, which can be easily detected. Hammoud et al.~\cite{Hammoud2021CheckYO} proposed to find the frequencies which are sensitive to the decision of the DNN models as the position to inject frequency trigger; however, the method needs a clean model well trained on the dataset prepared to poison and needs to mislabel the poisoned samples when training the backdoored model. 

In this paper, we propose an invisible backdoor attack taking advantage of the characteristics of the frequency domain, and we can inject a backdoor without mislabeling and access to the training process. The trigger overlaps the whole image which helps DNN models to learn the feature of the trigger, and this makes backdoor attacks successful in the scenario of clean-label supervised learning and no-label self-supervised learning because the model can perceive the features of original images and the trigger at the same time.

\subsection{Backdoor Defenses}
Due to the great harm caused by backdoor attacks, many defenses methods have been proposed in recent years, which are mainly divided into three types: the defenses against training data~\cite{Chen2019DetectingBA, Tran2018SpectralSI}, the defenses against model inputs~\cite{Chou2020SentiNetDL, Doan2020FebruusIP}, and the defenses against models~\cite{Wang2019NeuralCI, Guo2019TABORAH, Liu2019ABSSN}.

\vspace{2pt}\noindent\textbf{Defenses against training data.} 
Activations of the last hidden layer reflect high-level features used by the neural network to predict results. Chen et al. \cite{Chen2019DetectingBA} propose an activation clustering method, the activations of inputs belonging to the same label are separated and clustered by applying a k-means cluster with $k = 2$ after dimension reduction, and one of the clusters is poisoned samples. These poisoned samples are removed or relabeled with an accurate label. Tran et al. \cite{Tran2018SpectralSI} explore spectral signature, a technique based on robust statistic analysis, to identify and remove poisoned data samples from a potentially compromised training dataset. 

\vspace{2pt}\noindent\textbf{Defenses against model inputs.}  
SentiNet~\cite{Chou2020SentiNetDL} is proposed to detect a potential attack region of an image using Grad-CAM~\cite{selvaraju2017grad} developed for model interpretability and object detection. Then, the region can be checked manually to identify poisoned inputs, i.e., samples with a trigger patched. Februus~\cite{Doan2020FebruusIP} train a generative adversarial network (GAN) to repair images automatically after removing the suspicious areas masked by Grad-CAM.

\vspace{2pt}\noindent\textbf{Defenses against models.} 
Wang et al. \cite{Wang2019NeuralCI} firstly propose the Neural Cleanse to detect whether a DNN model has been backdoored or not prior to deployment by reversing the trigger. The performance of reversing triggers further improved in TABOR~\cite{Guo2019TABORAH} by utilizing various regularizations when solving optimizations.
ABS \cite{Liu2019ABSSN} examines whether a given DNN model is backdoored or not by analyzing inner neuron behaviors. In particular, a neuron that significantly contributes to a particular output label regardless of inputs is considered a compromised neuron. And then, ABS generates a trigger for the compromised neuron using the stimulation analysis and utilizes the performance of the trigger to confirm if the neuron is truly backdoored.


\section{Data Poisoning with Frequency Domain}

\begin{figure*}[!t]
\centering
\epsfig{figure=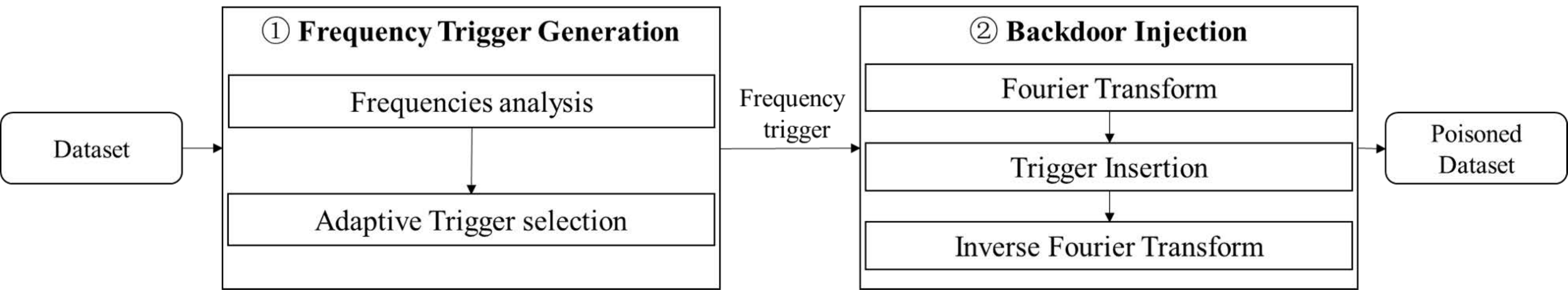, width=\textwidth} 
\caption{Overview of the frequency backdoor attack using data poisoning}
\label{fig:overview}
\end{figure*}



\subsection{Threat Model}
\label{4.1}



We consider an attacker aims to poison a dataset by patching an invisible trigger on the part of the samples so that there is no need to control the training process of the models, and any DNN model trained on the dataset containing backdoor samples, either by supervised learning and self-supervised learning, will be implanted with a backdoor. 
Specifically, attackers have three main goals: effectiveness, stealthiness, and robustness. \textit{Effectiveness} means that the model will have similar behavior to the benign model when processing samples without our trigger attached. However, it will misclassify samples patched with the trigger as a specific label with a high success rate. 
\textit{Stealthiness} requires that backdoor triggers are invisible to humans so that samples patched with triggers can pass the manual check. 
\textit{Robustness} represents two cases, one is that the trigger remains valid under some common defenses, and the other is that the performance of the model is significantly degraded if a defender tries to clear the trigger using some defenses against the input or dataset.


\subsection{Overview}
Figure~\ref{fig:overview} overviews our data poisoning attack approach, including two main components: Frequency Trigger Generation and Backdoor Injection. We aim to poison a dataset so that the DNN models trained on the dataset will be injected into the backdoor. Given an original dataset, we analyze the frequency distribution characteristics of the images in a preset target class and select a set of intensity values of proper frequencies in the frequency domain as triggers based on statistical features. To achieve our attack goals (see Section~\ref{4.1}), we design adaptive triggers in the frequency domain with the following metrics: invisible to human beings, patterns that overlap with large areas of the images in the spatial domain, no specific pattern in the spatial domain and still exist after common data preprocessing.
In the second phase, we transform the images in the target class from the spatial domain to the frequency domain using Fourier transform and inject our frequency trigger to generate a poisoned dataset.

\subsection{Trigger design}
As shown in~\cite{Xu2019TrainingBO, Xu2020FrequencyPF, Luo2021TheoryOT}, frequency domain information can be perceived by the DNN models, and its changes can affect all the pixels of the original trigger. This means that frequency triggers overlap the whole image and may take effect in the no-label or clean-label backdoor attack scenarios~\cite{Turner2018CleanLabelBA,Barni2019ANB,Liu2020ReflectionBA}. In addition, it is difficult for humans to perceive the changes in images if the intensity changes of specific frequencies are within a threshold, which means that the frequency domain may be a good way to insert triggers and utilize poisoning to achieve a backdoor attack. In this paper, we propose an approach for generating an adaptive frequency trigger of the following form:
\begin{equation}
\label{trigger}
F_T(u,v)=\begin{cases}
    0,&(u,v)\notin \nu_{T}\\
    I_T(u,v)-F_n(u,v),&(u,v)\in \nu_{T}
         \end{cases}
\end{equation} 
where $F_T(u,v)$ and $F_n(u,v)$ represent the intensity of the trigger and the original image at $(u,v)$ frequency, respectively, $\nu_{T}$ represents the frequency we choose to change the intensity, and $I_T(u,v)$ represents the target intensity we prepare to set for each selected frequency, and their selection follows the two objectives as below:
\begin{equation}
\label{robustness}
\begin{aligned}
\nu_{T}=&\mathop{minN}_{(u,v)}(Diff(F(u,v),F_{filter}(u,v))) \\
  &\cap \mathop{maxN}_{(u,v)}(Discrete(F(u,v)))
\end{aligned}
\end{equation}

\begin{equation}
\label{stealthiness}
\left | I_T(u,v)-F_n(u,v) \right | < \varepsilon
\end{equation}
where $F(u,v)$ and $F_{filter}(u,v)$ are the set of intensities at the frequency $(u,v)$ of all the images in $D_s$ and the images after filtering, $Diff$ is to calculate the average differences of the intensity at the frequency $(u,v)$ among all the images, and $Discrete$ is to calculate the dispersion of the intensities at the frequency $(u,v)$ among all the images, $\varepsilon$ is the threshold lower than which changes are not perceived by humans. Equation~\ref{robustness} selects the frequencies with strong robustness against filter and defense methods based on trigger pattern detection, and Equation~\ref{stealthiness} sets intensities with high stealthiness.
Finally, we patch our triggers into the image as follows: 
\begin{equation}
\label{image_with_patern}
F_n'(u,v)=\begin{cases}
         F_n(u,v),&(u,v)\notin \nu_{T}\\
         I_T(u,v),&(u,v)\in \nu_{T}
         \end{cases}
\end{equation}
where $F_n'(u,v)$ is the intensity of an poisoned image at the frequency $(u,v)$.

\begin{algorithm}[ht]
 	\caption{Adaptive trigger generation algorithm in the frequency domain}
 	\label{alg:adaptive-frequency-trigger-selection}
 	\begin{algorithmic}[1]
 	\REQUIRE $D_{t}$: the dataset with the target label in the frequency domain; $N$: the number of the images in $D_{t}$; $(C,W,H)$: the shape of each image; $\nu$: all the frequencies of images; $TopN$: the number of frequencies to select; $\varepsilon$: the threshold of intensities; 
 	\ENSURE $\nu_T$: the list of the frequencies selected; $I_T$: the list of the corresponding target intensities
 	
 	\STATE{$\nu = \{(1,1),(1,2),...,(W,H)\}$}
 	\STATE{$D_{t} = \{x^{n}\}, n \in [1,N]$}
 	\STATE{$x^{n} = \{F^{n}_{ch}(u,v)\}, (u,v) \in \nu, ch \in [1,C]$}
 	\STATE{$\nu_T = NULL, I_T = NULL$}

 	\STATE{$\nu_{candidate} = SelectFrequencyRobustToFilter(D_t)$}
 	\STATE{$\nu_{T} = SelectFrequencyDiscrete(D_t,\nu_{candidate})$}
 	
 	\STATE{$ I_T = GetMeanValue(D_{t}, \nu_T, ch), ch \in[1,C]$}
 	\STATE{$ I_T = SetValue(\nu_T, I_T, \varepsilon)$}
 	\STATE{\textbf{return} $\nu_T, I_T$}
 	
 	~\\

    \STATE{\textbf{Function} $SelectFrequencyRobustToFilter(D_t)$}
    \STATE{\quad$\{x^n_{filter}\} = Filter(\{x^n\}), n \in [1,N]$}
  	\STATE{\quad\textbf{for} $ch$ in $[1,C]$ \textbf{do}}
 	    \STATE{\quad\quad\textbf{for} $(u,v)$ in $\nu$ \textbf{do}}
	      \STATE{\quad\quad\quad$diff(u,v) = Diff(\{x^n\},\{x^n_{filter}\}), n \in [1,N]$}
 	    \STATE{\quad\quad\textbf{end for}}
 	    \STATE{\quad\quad$\nu_{ch} = \mathop{AscendingSort} \limits_{(u,v)}(diff(u,v))$}
 	\STATE{\quad\textbf{end for}}
 	\STATE{\quad$\nu_{temp} = {\{\nu_{ch}\}},ch\in[1,C]$}
 	\STATE{\quad$\nu_{candidate} = CommonTop(\nu_{temp},50)$}
 	\STATE{\quad\textbf{return} $\nu_{candidate}$}
 	\STATE{\textbf{end Function}}
 	
    ~\\

    \STATE{\textbf{Function} $SelectFrequencyDiscrete(D_t,\nu_{candidate})$}
  	\STATE{\quad\textbf{for} $ch$ in $[1,C]$ \textbf{do}}
 	    \STATE{\quad\quad\textbf{for} $(u,v)$ in $\nu_{candidate}$ \textbf{do}}
 	      \STATE{\quad\quad\quad$CoV(u,v) = CalCov(\{x^n\}), n \in [1,N]$}
 	    \STATE{\quad\quad\textbf{end for}}
 	    \STATE{\quad\quad$\nu_{ch} = \mathop{DescendingSort} \limits_{(u,v)}(CoV(u,v))$}
 	\STATE{\quad\textbf{end for}}
 	\STATE{\quad$\nu_{temp} = {\{\nu_{ch}\}},ch\in[1,C]$}
 	\STATE{\quad$\nu_T = CommonTop(\nu_{temp},TopN)$}
 	\STATE{\quad\textbf{return} $\nu_{T}$}
 	\STATE{\textbf{end Function}}
\end{algorithmic}
\end{algorithm}

Algorithm~\ref{alg:adaptive-frequency-trigger-selection} illustrate our adaptive trigger generation process. Given a dataset $D_t$ with $N$ images that have been transformed to the frequency domain, we first try to select appropriate frequencies as a candidate, which can be robust to filters (line 5). And then, we select frequencies that can make significant differences between trigger patterns on different images in the spatial domain from the candidate as they help to bypass the current backdoor defenses based on trigger pattern detection (line 6). Finally, we set proper intensities at each frequency selected to make our trigger invisible (lines 7-8). For each channel, we calculate the mean value of the intensities at each frequency in $\nu_T$ among all the images as the basic intensities (line 7), and then set values slightly more significant than the basic intensities but no greater than a threshold $\varepsilon$ (the threshold is obtained from manual adjustment and observation) as the target intensities $I_T$ (line 8). Up to now, we have finished the choice of our adaptive frequency trigger.

Especially, when selecting the frequencies candidate (lines 10-21), we first pass each image to the filter (line 11). For each channel, we calculate the average relative distance between the intensities at each frequency of original images and the images after filtering (lines 13-15) and sort the distances from smallest to largest to obtain the corresponding frequencies (line 16). Then, we select the top 50 frequencies ranked high on all three channels as the candidate $\nu_{candidate}$ (line 19).
After the candidate frequencies are selected, we use a similar method to select the target frequencies (lines 22-32) at which the Coefficient of Variation (CoV) of intensities are the largest (the more significant the value of CoV, the greater the dispersion of the data). The triggers in the spatial domain are formalized as follows:
\begin{equation}
\label{trigger_spatial}
\begin{aligned}
f_T(i,j)&=\sum_{u=0}^{M-1}\sum_{v=0}^{N-1}F_T(u,v)G(i,j,u,v)\\
    &=\sum_{(u,v)\in\nu}(I_T(u,v)-F_n(u,v))G(i,j,u,v)
\end{aligned}
\end{equation} 
Since $I_T(u,v)$ is a fixed value, to maximize the variance of trigger patterns (i.e., $f_T(i,j)$) on different images, we should select frequency at which intensities (i.e., $F_n(u,v)$) vary considerably among all the images.
We calculate the CoV for each frequency in the candidate set $\nu_{candidate}$ across all images (lines 24-26) and sort the CoV from the largest to the most minor (line 27) to get the corresponding frequencies. Finally, we pick a specified number (i.e., $TopN$) of frequencies that rank high on all three channels as the position of our frequency trigger $\nu_T$ (line 30).


\subsection{Backdoor Injection}
After the backdoor trigger is generated, we can then implant it into the target class images of the dataset to build the poisoned dataset.
Firstly, we transform the images in the target class from the spatial domain to the frequency domain using the Discrete Cosine Transform (DCT) after converting them into the YUV color space (see Section~\ref{2.1}). 
Secondly, we inject our trigger into the images presented in the frequency domain by adding the intensity values of the trigger to that of images at the corresponding frequencies. 
Finally, we transform images back to the spatial domain using the Inverse Discrete Cosine Transform (IDCT), and then we get a poisoned dataset.








\section{Experiment}
\subsection{Experiment Settings}

\vspace{3pt}\noindent\textbf{Dataset and model.} We utilize three popular datasets in our experiments, including CIFAR-10~\cite{krizhevsky2009learning}, STL-10~\cite{Coates2011AnAO} and GTSRB~\cite{stallkamp2011german}. And we choose ResNet-18~\cite{He2016DeepRL}, a classic DNN model to evaluate our backdoor attack.

\noindent$\bullet$\textit{CIFAR-10} is one of the most commonly used datasets in the task of object classification, which consists of 60,000 $32\times32\times3$ color images in 10 classes, with 50,000 training images and 10,000 testing images.

\noindent$\bullet$\textit{STL-10} is an image recognition dataset for developing unsupervised feature learning, deep learning, and self-taught learning algorithms. The STL-10 contains 5,000 labeled training images and 1,000 labeled testing images in 10 classes, and 100,000 unlabeled images. Each image has a size of $96\times96\times3$.

\noindent$\bullet$\textit{GTSRB} is widely used in the task of autonomous driving, which contains 43 classes of traffic signs, split into 39,209 training images and 12,630 test images. The size of each image is $32\times32\times3$.

\noindent\textbf{Scenarios.} We benchmark our attack approach under both self-supervised learning and supervised learning scenarios. 
For self-supervised learning, we use two widely used methods, SimCLR~\cite{Chen2020ASF} and MoCO v2~\cite{Chen2020ImprovedBW} to pre-train ResNet-18 as an image encoder on the poisoned CIFAR-10 dataset and then use the encoder for the downstream datasets CIFAR-10, STL-10 and GTSRB to train downstream classifiers. We train ResNet-18 on the poisoned CIFAR-10 and STL-10 datasets for supervised learning. Note that for self-supervised learning, we implement them based on the implementation shown in~\cite{simclr-pytorch} for SimCLR and \cite{moco-pytorch,Chen2020ImprovedBW} for MOCO V2, and use their default training parameters and data transformations. For supervised learning, we apply the commonly used parameters and data transformation in order not to disturb the normal training process.



\vspace{3pt}\noindent\textbf{Metrics.} We use the following four metrics to evaluate our approach:

\noindent$\bullet$\textit{Clean Data Performance (CDP)} evaluates the proportion of clean samples predicted as their ground-truth classes by classification models. 


\noindent$\bullet$\textit{Attack Success Rate (ASR)} is the fraction of the poisoned images (i.e., the images embedded with our trigger) that are predicted as the target label we specify by the backdoored DNN models.

\noindent$\bullet$\textit{Peak Signal-to-Noise Ratio (PSNR)} is the ratio between the maximum possible power of a signal and the power of corrupting noise that affects the fidelity of its representation. It's commonly used to quantify the quality of images after processing.

\noindent$\bullet$\textit{Structural Similarity (SSIM)} is used for measuring the similarity between two images. Compared to PSNR, SSIM is more in line with the perception of human beings because it incorporates crucial perceptual phenomena, including both luminance masking and contrast masking terms.

Using the metrics above, we regard our poisoning attack as a successful attack if it satisfies the following points: 1) The CDP of the backdoored model is similar to the original clean model, which means that our attack has little impact on the original performance. 2) The ASR of the backdoored model can meet our expectations, i.e., the model can achieve an ASR higher than 90\%. 3) The backdoored model's PSNR and SSIM are relatively high, which means that our triggers have no significant effect on the original images, i.e., it is difficult for humans to perceive the presence of our triggers. 4) The state-of-the-art defense work does not defend well against our attack, i.e., after applying the defenses, either the ASR of the backdoored model remains relatively high, or the CDP drops significantly together with the ASR.

\vspace{3pt}\noindent\textbf{Platform.} All our experiments are conducted on a server running 64-bit Ubuntu 20.04.3 system with Intel(R) Xeon(R) Platinum 8268 CPU @ 2.90GHz, 188GB memory, 20TB hard drive, and one Nvidia GeForce RTX 3090 GPUs with 24GB memory.

\subsection{Effectiveness}

\begin{table}[]
\caption{Baseline of Clean Self-Supervised Learning Model}
\label{tab:effectiveness-of-Baseline-SSL}
\begin{tabular}{c|c|c|c|c}
\hline
\textbf{Method}          &  \textbf{\begin{tabular}[c]{@{}c@{}}Downstream\\ Dataset\end{tabular}} & \textbf{CDP} & \textbf{\begin{tabular}[c]{@{}c@{}}Target\\ Label\end{tabular}} & \textbf{ASR} \\ \hline \hline
\multirow{3}{*}{SimCLR}                                               & CIFAR-10  &      86.53\%   &       automobile                                                        &   9.31\%            \\
                                    & GTSRB      &     91.38\%                &            speed limit 30                                         & 12.6\%             \\
                               & STL-10      &    77.21\%                &              car                     & 2.70\%             \\ \hline
\multirow{3}{*}{MoCo v2}                & CIFAR-10     &       84.03\%          &         automobile        &      9.29\%        \\
                                     & GTSRB       &      89.64\%                       &              speed limit 100            &       8.38\%       \\
                    & STL-10           &     72.65\%            &          car                     &     3.11\%         \\ \hline
\end{tabular}
\end{table}



\begin{table}[]
\caption{Baseline of Clean Supervised Learning Model}
\label{tab:effectiveness-of-Baseline-SL}
\begin{tabular}{c|c|c|c}
\hline
\textbf{Model} & \textbf{Dataset} &  \textbf{CDP} & \textbf{ASR} \\ \hline \hline
\multirow{2}{*}{Resnet18}      & CIFAR-10          &        91.06\%        &     9.96\%        \\ \cline{2-4} 
     & STL-10        &     74.42\%        &  5.46\%     \\\hline
\end{tabular}

\end{table}

\begin{table}[]
\caption{Effectiveness of Backdoor Attack on Self-Supervised Learning}
\label{tab:effectiveness-of-backdoored-SSL}
\begin{tabular}{c|c|c|c}
\hline
\textbf{Method}          &  \textbf{\begin{tabular}[c]{@{}c@{}}Downstream\\ Dataset\end{tabular}} & \textbf{CDP} & \textbf{ASR} \\ \hline \hline
\multirow{3}{*}{SimCLR}    & CIFAR-10                                                 &      85.48\%(-1.05\%)           &      92.08\%        \\
                                & GTSRB                                           &      89.67\%(-1.71\%)              &    95.57\%          \\
                    & STL-10                                                &     76.76\%(-0.45)        &     99.68\%         \\ \hline
\multirow{3}{*}{MoCo v2}    & CIFAR-10                                       &     82.24\%(-1.79\%)            &    91.55\%          \\
           & GTSRB                                          &      87.90\%(-1.74\%)      &      94.60\%        \\
                      & STL-10                                             &     71.83\%(-0.82)                 &      99.78\%        \\ \hline
\end{tabular}
\end{table}


\begin{table}[]
\caption{Effectiveness of Backdoor Attack on Supervised Learning}
\label{tab:effectiveness-of-backdoored-SL}
\begin{tabular}{c|c|c|c}
\hline
\textbf{Model} & \textbf{Dataset} &  \textbf{CDP} & \textbf{ASR} \\ \hline \hline
\multirow{2}{*}{Resnet18}      & CIFAR-10          &        90.63\%(-0.43\%)            &     95.56\%       \\ \cline{2-4} 
     & STL-10        &     72.51\%(-1.91\%)               &     92.45\%  \\\hline
\end{tabular}
\end{table}

\subsubsection{Baseline Performance}
\ 
\newline
For the scenario of self-supervised learning, we train clean DNN encoders (i.e., ResNet-18) on CIFAR-10 using two self-supervised methods (i.e., SimCLR and MoCo V2). And then use the encoder to train downstream classifiers on CIFAR-10, STL-10, and GTSRB tasks. We evaluate their performance in several aspects as the baseline in Table~\ref{tab:effectiveness-of-Baseline-SSL}. Moreover, for supervised learning, we train clean ResNet-18 models on CIFAR-10 and STL-10, and the effectiveness is shown in Table~\ref{tab:effectiveness-of-Baseline-SL}.
As we can see from the two tables, the clean models achieve similar performance as those shown in~\cite{Chen2020ASF,Chen2020ImprovedBW, He2016DeepRL}. In addition, the ASR tested on the poisoned dataset is low for all the clean models proves that our trigger has little influence on the decision of main task.

\subsubsection{Backdoor Performance}
\ 
\newline
We evaluate our backdoor attack on the same tasks as those shown in the baseline. We generate triggers using our proposed adaptive algorithm~\ref{alg:adaptive-frequency-trigger-selection}. For example, we select ``automobile'' as the poisoned label in CIFAR-10 dataset, and the trigger generated is composed of frequencies (1,10), (1,9) and (0,10), at which the intensities on three channels are set to (70,70,80), (65,65,65) and (65,65,65) respectively.
The backdoor performance on self-supervised learning and supervised learning are shown in Table~\ref{tab:effectiveness-of-backdoored-SSL} and Table~\ref{tab:effectiveness-of-backdoored-SL} respectively. 
The experimental result illustrates that our method achieves satisfactory ASR (i.e., above 90\%) on all the tasks with an acceptable reduction of the performance on the clean samples (i.e., lower than 2\%). The result confirms that DNNs can learn the trigger feature that can cause global perturbations to the original images in the clean-label and no-label scenarios. Specifically, in the scenario of self-supervised learning, the features of images patched with the trigger can be clustered with the features of original images. 
In other words, our trigger's feature can be regarded as one of the features of the images with the target category.

Note that in the experiment of self-supervised learning, we do not know which label in the downstream task will be poisoned before the model is deployed, only after the model is put into use, we use our trigger to activate the backdoor, and we can know the corresponding target label (the target label in our experiments is shown in Table~\ref{tab:effectiveness-of-Baseline-SSL}), which means we achieve an untargeted backdoor attack here. However, we find something interesting when transferring the encoder pre-trained on CIFAR-10 to downstream task STL-10, the target label is the same as the poisoned label set when training the poisoned encoder. This is because the features of the trigger are bounded to the features of the pretraining samples with the poisoned label, which can also be bounded to the downstream samples with similar features after transferring. In this way, we can also achieve a targeted backdoor attack as long as the downstream dataset and our poisoned dataset have overlapping categories.

\begin{table}[]
\caption{The invisibility of triggers of different stealthy backdoor attack}
\label{tab:invisibility}
\begin{tabular}{c|c|c}
\hline
\textbf{attack method} & \textbf{PSNR} &  \textbf{SSIM} \\ \hline \hline
  Blend\cite{Chen2017TargetedBA}        &      19.18          &     0.7921      \\\hline
  SIG\cite{Barni2019ANB}    &     25.12          &        0.8988    \\ \hline
  REFOOL\cite{Liu2020ReflectionBA}         &      16.59         &      0.7701       \\\hline
  Ours      &     24.11       &        0.9024  \\\hline
\end{tabular}
\end{table}

\subsubsection{Invisibility}
\ 
\newline
We evaluate the invisibility of our trigger by calculating the average PSNR and SSIM of CIFAR-10 and STL-10 images patched with the trigger. The PSNR on CIFAR-10 and STL-10 datasets is 24.11 and 26.94, respectively, and the SSIM is 0.9024 and 0.9044.
Moreover, we compare the invisibility with state-of-the-art backdoor attacks using the CIFAR-10 images patched with different triggers. The result in Table~\ref{tab:invisibility} shows that our trigger is superior to current work due to the characteristics of the frequency domain.

\begin{table}[]
\caption{Resistance to different filters}
\label{tab:resistance-to-filtering}
\begin{tabular}{c|c|c}
\hline
\textbf{Filter} & \textbf{CDP}& \textbf{ASR} \\ \hline \hline
Gaussian Filter                 &     57.26\%                &       86.12\%     \\ \hline
Mean Filter                &      51.84\%           &     79.23\%         \\ \hline
Median Filter                 &     76.50\%            &      88.66\%               \\ \hline
SVD Filter                 &     74.33\%                 &     93.52\%              \\ \hline
\end{tabular}
\end{table}

\begin{table*}[]
\caption{Evaluation of different intensities settings.}
\label{tab:effectiveness-of-intensities}
\begin{threeparttable}
\begin{tabular}{c|c|c|c|c|c|c}
\hline
  \textbf{No.}& \textbf{itensities} & \textbf{PSNR} & \textbf{SSIM} & \textbf{CDP}  & \textbf{ASR} & \textbf{\begin{tabular}[c]{@{}c@{}}ASR after Filter\end{tabular}} \\ \hline \hline
 1  &   (60,60,70),(55,55,55),(55,55,55)                &      25.03         &     0.9122          &    91.00\%          &     87.84\%           &        79.80\%        \\ \hline
   2(ours)  &   (70,70,80),(65,65,65),(65,65,65)                &      24.11         &     0.9024          &    90.63\%          &     91.00\%           &        86.12\%        \\\hline
   3  &   (80,80,90),(75,75,75),(75,75,75)                &      23.23         &     0.8911         &    90.77\%          &     91.62\%           &        78.80\%        \\ \hline

\end{tabular}
\footnotesize
Note: The intensities increase from No.1 to No.3.
\end{threeparttable}
\end{table*}

\begin{table*}[]
\caption{Evaluation of different frequencies settings}
\label{tab:effectiveness-of-frequencies}
\begin{threeparttable}
\begin{tabular}{c|c|c|c|c|c|c|c}
\hline
  \textbf{No.} &\textbf{frequencies} & \textbf{itensities} & \textbf{PSNR} & \textbf{SSIM} & \textbf{CDP}  & \textbf{ASR} & \textbf{\begin{tabular}[c]{@{}c@{}}ASR after Filter\end{tabular}} \\ \hline \hline
 1 & (28,0),(30,0),(31,0)                &     (40,35,30),(25,25,25),(25,25,25)          &     31.20          &     0.9530         &      90.29\%          &   97.15\%  &    4.32\%   \\ \hline
   2(ours) &  (1,10),(1,9),(0,10)  &   (70,70,80),(65,65,65),(65,65,65)                &      24.11         &     0.9024          &    90.63\%          &     91.00\%           &        86.12\%  \\ \hline
                3 &       (1,7),(3,6),(5,3)                 &   (95,105,130),(85,85,85),(85,85,85)            &  21.57             &   0.8624           &   89.00\%     &  88.18\%    &   84.68\%     \\ \hline

\end{tabular}
\footnotesize
Note: The frequencies decrease from No.1 to No.3, and the corresponding intensities increase to ensure ASR is larger than 85\% and CDP is around 90\%.
\end{threeparttable}
\end{table*}



\subsection{Impacts of Intensities and Frequencies}

\noindent\textbf{Impacts of intensities.}
To evaluate the influence of intensities, we choose triggers with lower and higher intensities than our trigger, and then inject them into the CIFAR-10 dataset. As shown in Table~\ref{tab:effectiveness-of-intensities}, trigger with larger intensities is easier to achieve backdoor attack (i.e., the ASR is larger), but it losses its stealthiness at the same time (i.e., the PSNR and SSIM get lower).

\noindent\textbf{Impacts of frequencies.} For frequencies, we select triggers with higher and lower frequencies and set proper intensities for them to ensure the ASR is larger than 85\% and CDP is around 90\%. As shown in Table~\ref{tab:effectiveness-of-frequencies}, trigger with higher frequencies is easier to be perceived by DNN models and a little stealthier than the trigger selected by our proposed method, but it is more likely to be filtered out by the filters (No.1). On the contrary, trigger with lower frequencies is challenging to learn by DNN models, so we need to set slightly higher intensities, which may cause visible changes to the images in the spatial domain and affect the performance of the clean dataset (No.3).

\subsection{Resistance}
We choose four widely used defenses in different types that are most relevant to our attack to evaluate the robustness of our attack, including the detection of training data (i.e., Activation Clustering), the preprocessing of inputs (i.e., Filtering), the detection of inputs (i.e., SentiNet) and the detection of models (i.e., Neural Cleanse).

\subsubsection{Resistance tor Activation Clustering}
\ 
\newline
Activation Clustering is an outlier detection method commonly used to detect poisoned data since a poisoned input may make the distribution of activations different from a clean input, which amplifies signals critical for classification to be detected by clustering. 
However, Activation Clustering can only operate on the samples with target labels, so it is not suitable in the scenario of self-supervised learning. 
We evaluate the backdoored ResNet-18 model trained on our poisoned CIFAR-10 dataset for supervised learning. For the samples with the target label, the FPR (i.e., the ratio of clean samples that Activation Clustering misclassifies to be poisoned samples) is 100.00\%, and the FNR (i.e., the ratio of poisoned samples that the method regards as benign samples) is 56.64\%. The results show that although many poisoned samples are successfully detected, all the benign samples are also classified to be malicious, which leads to a significant performance drop in the DNN models trained on the dataset processed, and the backdoor can still be injected because the samples with the target label left are all poisoned. In addition, nearly half of the poisoned samples are classified as benign, which means that the activation difference between benign and poisoned samples is not significant, implying the high stealthiness of our backdoor attack.

\subsubsection{Resistance to Filtering}
\ 
\newline
We evaluate the robustness of our attack to the filters on the backdoored model ResNet-18 that trained on poisoned dataset CIFAR-10 using supervised learning. Before feeding the test samples into the model for prediction, we first pass them through four types of filters (i.e., Gaussian, average, median, and SVD filters). Note that the SVD filter in the frequency domain is used to filter out singular frequencies by analyzing the frequency distribution.
The result in Table~\ref{tab:resistance-to-filtering} proves that the ASR of our backdoor attack remains high after filtering, but the performance on the clean data drops significantly, which means our backdoor attack is robust to the processing of filters.

\begin{figure}[!t]
\centering
\epsfig{figure=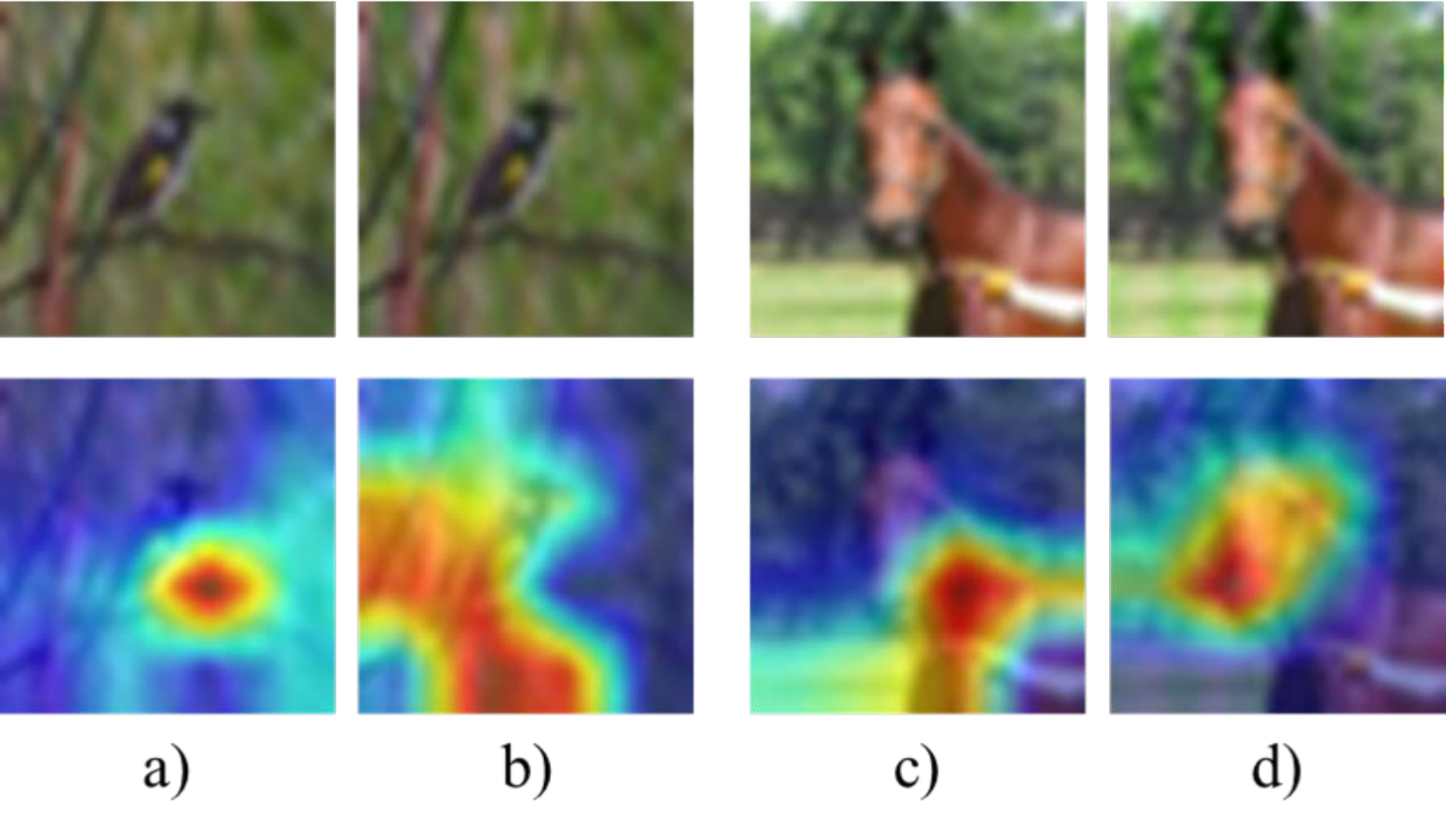, width=0.45\textwidth} 
\caption{Critical Regions Identified by SentiNet. Column a) and c) are the results of clean images, and column b) and d) are the results of the corresponding poisoned images}
\label{fig:Grad_CAM}
\vspace{-10pt}
\end{figure}

\subsubsection{Resistance to SentiNet}
\ 
\newline
SentiNet is used to detect a potential attack region of an image using Grad-CAM developed for model interpretability and object detection. And then the region can be checked manually to identify poisoned inputs, i.e., samples with a trigger patched. 
We apply SentiNet to the backdoored ResNet-18 models trained by self-supervised learning using SimCLR to see if the trigger patched on the poisoned CIFAR-10 dataset can be detected. Figure~\ref{fig:Grad_CAM} shows the regions identified by SentiNet on several randomly selected samples, where column a) and c) are the results of clean images, column b) and d) are the results of the corresponding poisoned images, from which we can see that SentiNet can only identify a partial area of an image as a critical region, however, our trigger overlaps the whole image. In addition, our trigger truly changes the focus of the model on the image compared to the clean image, but the regions identified on poisoned images are still the significant regions of images that lead DNN models to identify the original images, which means that humans cannot detect our trigger using such method.

\subsubsection{Resistance to Neural Cleanse}
\begin{figure}[!t]
\centering
\epsfig{figure=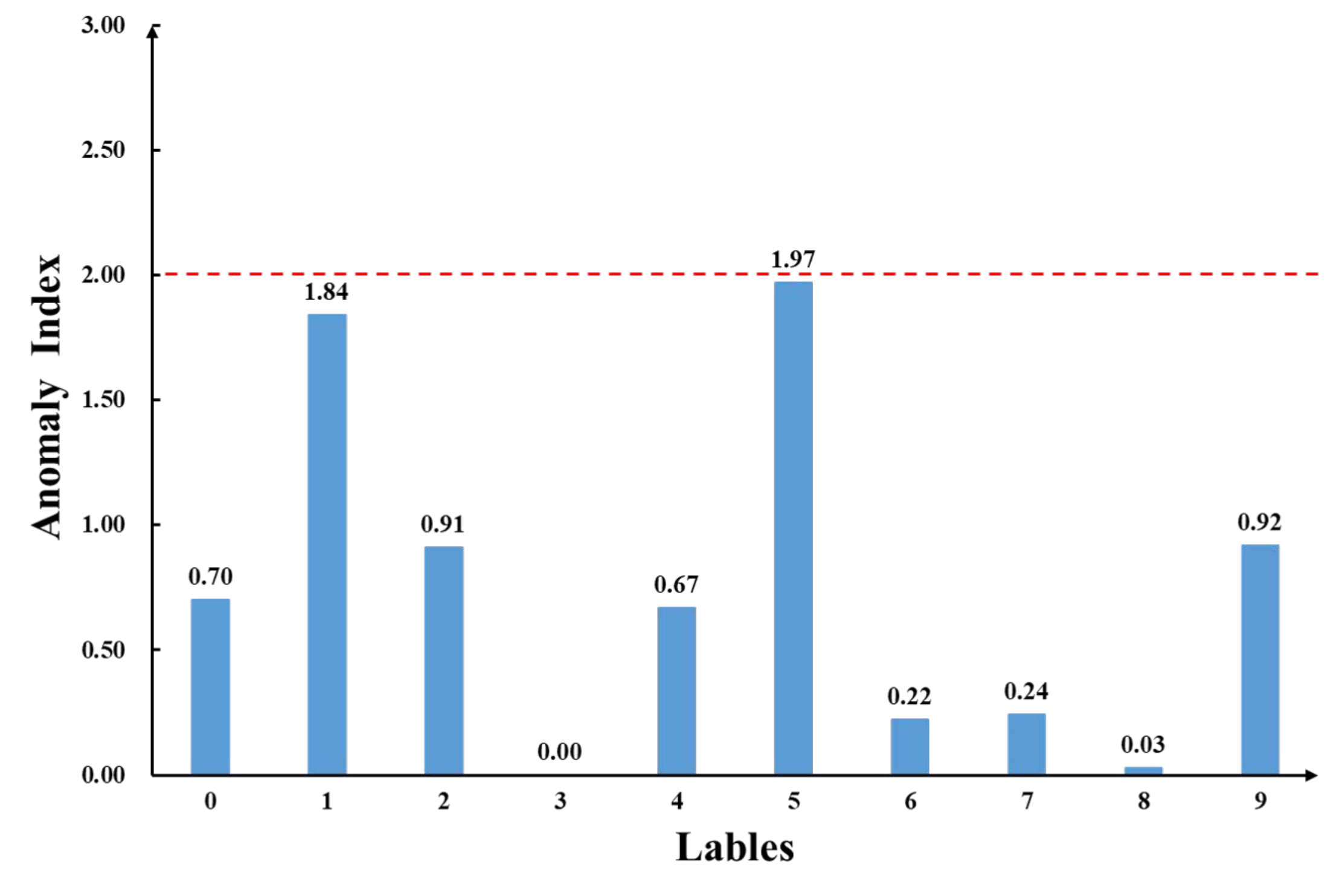, width=0.47\textwidth} 
\caption{Critical regions identified by SentiNet}
\label{fig:NC}
\vspace{-10pt}
\end{figure}

\ 
\newline
Neural Cleanse is the most widely used defense against backdoor attacks by detecting the DNN models. For each label, Neural Cleanse tries to find a potential minimal trigger that can misclassify all samples into this target label, and then use an outlier detection algorithm to choose the trigger that is significantly smaller than the others as the real trigger (i.e., the anomaly index is larger than 2), and the corresponding label is the target label of the backdoor attack.  
We apply Neural Cleanse to detect our backdoored ResNet-18 model trained on CIFAR-10 using SimCLR. The anomaly index of each label is shown in Figure~\ref{fig:NC}, which shows that none of the anomaly indexes are larger than 2, proving that Neural Cleanse cannot find our frequency trigger.

\section{Discussion}
In this paper, we propose an invisible backdoor attack using data poisoning. Due to the high stealthiness of our frequency trigger, it can also be used to watermark datasets or DNN models. Firstly, we can release a dataset in which all the samples are patched with our invisible trigger, and our backdoored model trained before will output the target label on any samples, but a new model trained on them will not. 
Secondly, we can release our backdoored model and then use our trigger to verify the ownership of the model.
Besides, since the frequency domain is derived from signal processing, our idea of the backdoor attack (i.e., based on some statistic analysis in the frequency domain) may be able to migrate to the speech domain as well.

\section{Conclusion}
In this paper, we propose a generalized backdoor attack approach based on the frequency domain, where the DNN models trained on our poisoned dataset will be implanted with a backdoor without mislabeling the poisoned samples and accessing the training process. We evaluate our approach in the no-label and clean-label cases on popular benchmark datasets with self-supervised learning and supervised learning and test the effectiveness of the current defenses against the backdoor attack. The results show that our approach can achieve a high attack success rate on all the tasks without significant performance downgrade on the main tasks, and also, it can evade the commonly used defense methods.

\bibliographystyle{plain}
\bibliography{reference}

\begin{thebibliography}{10}

\bibitem{ap}
Baidu apollo team (2017), apollo: Open source autonomous driving.
\newblock \url{https://github.com/ApolloAuto/apollo}.

\bibitem{BAI202165}
Zhongxin Bai and Xiao-Lei Zhang.
\newblock Speaker recognition based on deep learning: An overview.
\newblock {\em Neural Networks}, 140:65--99, 2021.

\bibitem{Barni2019ANB}
Mauro Barni, Kassem Kallas, and Benedetta Tondi.
\newblock A new backdoor attack in cnns by training set corruption without
  label poisoning.
\newblock {\em 2019 IEEE International Conference on Image Processing (ICIP)},
  pages 101--105, 2019.

\bibitem{Carlini2021PoisoningAB}
Nicholas Carlini and A.~Terzis.
\newblock Poisoning and backdooring contrastive learning.
\newblock {\em ArXiv}, abs/2106.09667, 2021.

\bibitem{Chen2019DetectingBA}
Bryant Chen, Wilka Carvalho, Nathalie Baracaldo, Heiko Ludwig, Ben Edwards,
  Taesung Lee, Ian Molloy, and B.~Srivastava.
\newblock Detecting backdoor attacks on deep neural networks by activation
  clustering.
\newblock {\em ArXiv}, abs/1811.03728, 2019.

\bibitem{Chen2020ASF}
Ting Chen, Simon Kornblith, Mohammad Norouzi, and Geoffrey~E. Hinton.
\newblock A simple framework for contrastive learning of visual
  representations.
\newblock {\em ArXiv}, abs/2002.05709, 2020.

\bibitem{Chen2020ImprovedBW}
Xinlei Chen, Haoqi Fan, Ross~B. Girshick, and Kaiming He.
\newblock Improved baselines with momentum contrastive learning.
\newblock {\em ArXiv}, abs/2003.04297, 2020.

\bibitem{Chen2017TargetedBA}
Xinyun Chen, Chang Liu, Bo~Li, Kimberly Lu, and Dawn~Xiaodong Song.
\newblock Targeted backdoor attacks on deep learning systems using data
  poisoning.
\newblock {\em ArXiv}, abs/1712.05526, 2017.

\bibitem{Chou2020SentiNetDL}
Edward Chou, Florian Tram{\`e}r, and Giancarlo Pellegrino.
\newblock Sentinet: Detecting localized universal attacks against deep learning
  systems.
\newblock {\em 2020 IEEE Security and Privacy Workshops (SPW)}, pages 48--54,
  2020.

\bibitem{Coates2011AnAO}
Adam Coates, A.~Ng, and Honglak Lee.
\newblock An analysis of single-layer networks in unsupervised feature
  learning.
\newblock In {\em AISTATS}, 2011.

\bibitem{Doan2020FebruusIP}
Bao~Gia Doan, Ehsan Abbasnejad, and Damith~Chinthana Ranasinghe.
\newblock Februus: Input purification defense against trojan attacks on deep
  neural network systems.
\newblock {\em Annual Computer Security Applications Conference}, 2020.

\bibitem{Gardner2017AllenNLP}
Matt Gardner, Joel Grus, Mark Neumann, Oyvind Tafjord, Pradeep Dasigi,
  Nelson~F. Liu, Matthew Peters, Michael Schmitz, and Luke~S. Zettlemoyer.
\newblock Allennlp: A deep semantic natural language processing platform.
\newblock 2017.

\bibitem{Gu2019BadNetsEB}
Tianyu Gu, Kang Liu, Brendan Dolan-Gavitt, and Siddharth Garg.
\newblock Badnets: Evaluating backdooring attacks on deep neural networks.
\newblock {\em IEEE Access}, 7:47230--47244, 2019.

\bibitem{Gueguen2018FasterNN}
Lionel Gueguen, Alexander Sergeev, Benjamin~J. Kadlec, Rosanne Liu, and Jason
  Yosinski.
\newblock Faster neural networks straight from jpeg.
\newblock In {\em NeurIPS}, 2018.

\bibitem{Guo2019TABORAH}
Wenbo Guo, Lun Wang, Xinyu Xing, Min Du, and Dawn~Xiaodong Song.
\newblock Tabor: A highly accurate approach to inspecting and restoring trojan
  backdoors in ai systems.
\newblock {\em ArXiv}, abs/1908.01763, 2019.

\bibitem{Hammoud2021CheckYO}
Hasan Abed Al~Kader Hammoud and Bernard Ghanem.
\newblock Check your other door! creating backdoor attacks in the frequency
  domain.
\newblock 2021.

\bibitem{He2016DeepRL}
Kaiming He, X.~Zhang, Shaoqing Ren, and Jian Sun.
\newblock Deep residual learning for image recognition.
\newblock {\em 2016 IEEE Conference on Computer Vision and Pattern Recognition
  (CVPR)}, pages 770--778, 2016.

\bibitem{Jia2021BadEncoderBA}
Jinyuan Jia, Yupei Liu, and Neil~Zhenqiang Gong.
\newblock Badencoder: Backdoor attacks to pre-trained encoders in
  self-supervised learning.
\newblock {\em ArXiv}, abs/2108.00352, 2021.

\bibitem{Kipf2017SemiSupervisedCW}
Thomas Kipf and Max Welling.
\newblock Semi-supervised classification with graph convolutional networks.
\newblock {\em ArXiv}, abs/1609.02907, 2017.

\bibitem{krizhevsky2009learning}
Alex Krizhevsky, Geoffrey Hinton, et~al.
\newblock Learning multiple layers of features from tiny images.
\newblock 2009.

\bibitem{lecun1998gradient}
Yann LeCun, L{\'e}on Bottou, Yoshua Bengio, and Patrick Haffner.
\newblock Gradient-based learning applied to document recognition.
\newblock {\em Proceedings of the IEEE}, 86(11):2278--2324, 1998.

\bibitem{Liu2019ABSSN}
Yingqi Liu, Wen-Chuan Lee, Guanhong Tao, Shiqing Ma, Yousra Aafer, and
  X.~Zhang.
\newblock Abs: Scanning neural networks for back-doors by artificial brain
  stimulation.
\newblock {\em Proceedings of the 2019 ACM SIGSAC Conference on Computer and
  Communications Security}, 2019.

\bibitem{Liu2018TrojaningAO}
Yingqi Liu, Shiqing Ma, Yousra Aafer, Wen-Chuan Lee, Juan Zhai, Weihang Wang,
  and X.~Zhang.
\newblock Trojaning attack on neural networks.
\newblock In {\em NDSS}, 2018.

\bibitem{Liu2020ReflectionBA}
Yunfei Liu, Xingjun Ma, James Bailey, and Feng Lu.
\newblock Reflection backdoor: A natural backdoor attack on deep neural
  networks.
\newblock In {\em ECCV}, 2020.

\bibitem{Luo2021SelfSupervisedPM}
Chenxu Luo, Xiaodong Yang, and Alan~Loddon Yuille.
\newblock Self-supervised pillar motion learning for autonomous driving.
\newblock {\em 2021 IEEE/CVF Conference on Computer Vision and Pattern
  Recognition (CVPR)}, pages 3182--3191, 2021.

\bibitem{Luo2021TheoryOT}
Tao Luo, Zheng Ma, Zhi-Qin~John Xu, and Yaoyu Zhang.
\newblock Theory of the frequency principle for general deep neural networks.
\newblock {\em ArXiv}, abs/1906.09235, 2021.

\bibitem{Melamud2016context2vecLG}
Oren Melamud, Jacob Goldberger, and Ido Dagan.
\newblock context2vec: Learning generic context embedding with bidirectional
  lstm.
\newblock In {\em CoNLL}, 2016.

\bibitem{Mikolov2013DistributedRO}
Tomas Mikolov, Ilya Sutskever, Kai Chen, Gregory~S. Corrado, and Jeffrey Dean.
\newblock Distributed representations of words and phrases and their
  compositionality.
\newblock In {\em NIPS}, 2013.

\bibitem{parkhi2015deep}
Omkar~M Parkhi, Andrea Vedaldi, and Andrew Zisserman.
\newblock Deep face recognition.
\newblock 2015.

\bibitem{Podpora2014YUVVR}
Michal Podpora, Grzegorz~Pawel Korbas, and Aleksandra Kawala-Janik.
\newblock Yuv vs rgb-choosing a color space for human-machine interaction.
\newblock In {\em FedCSIS}, 2014.

\bibitem{Radford2021LearningTV}
Alec Radford, Jong~Wook Kim, Chris Hallacy, Aditya Ramesh, Gabriel Goh,
  Sandhini Agarwal, Girish Sastry, Amanda Askell, Pamela Mishkin, Jack Clark,
  Gretchen Krueger, and Ilya Sutskever.
\newblock Learning transferable visual models from natural language
  supervision.
\newblock In {\em ICML}, 2021.

\bibitem{2018FDA}
M.~Ratner.
\newblock Fda backs clinician-free ai imaging diagnostic tools.
\newblock {\em Nature Biotechnology}, 36(8):673--674, 2018.

\bibitem{Redmon2016YouOL}
Joseph Redmon, Santosh~Kumar Divvala, Ross~B. Girshick, and Ali Farhadi.
\newblock You only look once: Unified, real-time object detection.
\newblock {\em 2016 IEEE Conference on Computer Vision and Pattern Recognition
  (CVPR)}, pages 779--788, 2016.

\bibitem{moco-pytorch}
Hao Ren.
\newblock Moco.
\newblock \url{https://github.com/leftthomas/MoCo}, 2020.

\bibitem{simclr-pytorch}
Hao Ren.
\newblock Simclr.
\newblock \url{https://github.com/leftthomas/SimCLR}, 2020.

\bibitem{Ren2015FasterRT}
Shaoqing Ren, Kaiming He, Ross~B. Girshick, and Jian Sun.
\newblock Faster r-cnn: Towards real-time object detection with region proposal
  networks.
\newblock {\em IEEE Transactions on Pattern Analysis and Machine Intelligence},
  39:1137--1149, 2015.

\bibitem{Saha2021BackdoorAO}
Aniruddha Saha, Ajinkya Tejankar, Soroush~Abbasi Koohpayegani, and Hamed
  Pirsiavash.
\newblock Backdoor attacks on self-supervised learning.
\newblock {\em ArXiv}, abs/2105.10123, 2021.

\bibitem{Schroff2015FaceNetAU}
Florian Schroff, Dmitry Kalenichenko, and James Philbin.
\newblock Facenet: A unified embedding for face recognition and clustering.
\newblock {\em 2015 IEEE Conference on Computer Vision and Pattern Recognition
  (CVPR)}, pages 815--823, 2015.

\bibitem{selvaraju2017grad}
Ramprasaath~R Selvaraju, Michael Cogswell, Abhishek Das, Ramakrishna Vedantam,
  Devi Parikh, and Dhruv Batra.
\newblock Grad-cam: Visual explanations from deep networks via gradient-based
  localization.
\newblock In {\em Proceedings of the IEEE international conference on computer
  vision}, pages 618--626, 2017.

\bibitem{Sonka1993ImagePA}
Milan Sonka, V{\'a}clav Hlav{\'a}c, and Roger~D. Boyle.
\newblock Image processing, analysis and machine vision.
\newblock In {\em Springer US}, 1993.

\bibitem{stallkamp2011german}
Johannes Stallkamp, Marc Schlipsing, Jan Salmen, and Christian Igel.
\newblock The german traffic sign recognition benchmark: a multi-class
  classification competition.
\newblock In {\em The 2011 international joint conference on neural networks},
  pages 1453--1460. IEEE, 2011.

\bibitem{2013Intriguing}
C.~Szegedy, W.~Zaremba, I.~Sutskever, J.~Bruna, D.~Erhan, I.~Goodfellow, and
  R.~Fergus.
\newblock Intriguing properties of neural networks.
\newblock {\em Computer Science}, 2013.

\bibitem{Tao2021SelfsupervisedSR}
Ruijie Tao, Kong-Aik Lee, Rohan~Kumar Das, Ville Hautamaki, and Haizhou Li.
\newblock Self-supervised speaker recognition with loss-gated learning.
\newblock 2021.

\bibitem{Tran2018SpectralSI}
Brandon Tran, Jerry Li, and Aleksander Madry.
\newblock Spectral signatures in backdoor attacks.
\newblock In {\em NeurIPS}, 2018.

\bibitem{Turner2018CleanLabelBA}
Alexander Turner, Dimitris Tsipras, and Aleksander Madry.
\newblock Clean-label backdoor attacks.
\newblock 2018.

\bibitem{Wang2019NeuralCI}
Bolun Wang, Yuanshun Yao, Shawn Shan, Huiying Li, Bimal Viswanath, Haitao
  Zheng, and Ben~Y. Zhao.
\newblock Neural cleanse: Identifying and mitigating backdoor attacks in neural
  networks.
\newblock {\em 2019 IEEE Symposium on Security and Privacy (SP)}, pages
  707--723, 2019.

\bibitem{Wang2020HighFrequencyCH}
Haohan Wang, Xindi Wu, Pengcheng Yin, and Eric~P. Xing.
\newblock High-frequency component helps explain the generalization of
  convolutional neural networks.
\newblock {\em 2020 IEEE/CVF Conference on Computer Vision and Pattern
  Recognition (CVPR)}, pages 8681--8691, 2020.

\bibitem{Wang2021BackdoorAT}
Tong Wang, Yuan Yao, Feng Xu, Shengwei An, Hanghang Tong, and Ting Wang.
\newblock Backdoor attack through frequency domain.
\newblock {\em ArXiv}, abs/2111.10991, 2021.

\bibitem{Wang2019DynamicGC}
Yue Wang, Yongbin Sun, Ziwei Liu, Sanjay~E. Sarma, Michael~M. Bronstein, and
  Justin~M. Solomon.
\newblock Dynamic graph cnn for learning on point clouds.
\newblock {\em ACM Transactions on Graphics (TOG)}, 38:1 -- 12, 2019.

\bibitem{Wiles2018SelfsupervisedLO}
Olivia Wiles, A.~Sophia Koepke, and Andrew Zisserman.
\newblock Self-supervised learning of a facial attribute embedding from video.
\newblock In {\em BMVC}, 2018.

\bibitem{Xiao2018GeneratingAE}
Chaowei Xiao, Bo~Li, Jun-Yan Zhu, Warren He, M.~Liu, and Dawn~Xiaodong Song.
\newblock Generating adversarial examples with adversarial networks.
\newblock {\em ArXiv}, abs/1801.02610, 2018.

\bibitem{Xu2020LearningIT}
Kai Xu, Minghai Qin, Fei Sun, Yuhao Wang, Yen kuang Chen, and Fengbo Ren.
\newblock Learning in the frequency domain.
\newblock {\em 2020 IEEE/CVF Conference on Computer Vision and Pattern
  Recognition (CVPR)}, pages 1737--1746, 2020.

\bibitem{Xu2020FrequencyPF}
Zhi-Qin~John Xu, Yaoyu Zhang, Tao Luo, Yan Xiao, and Zheng Ma.
\newblock Frequency principle: Fourier analysis sheds light on deep neural
  networks.
\newblock {\em ArXiv}, abs/1901.06523, 2020.

\bibitem{Xu2019TrainingBO}
Zhi-Qin~John Xu, Yaoyu Zhang, and Yan Xiao.
\newblock Training behavior of deep neural network in frequency domain.
\newblock In {\em ICONIP}, 2019.

\bibitem{Yin2019AFP}
Dong Yin, Raphael~Gontijo Lopes, Jonathon Shlens, Ekin~Dogus Cubuk, and Justin
  Gilmer.
\newblock A fourier perspective on model robustness in computer vision.
\newblock {\em ArXiv}, abs/1906.08988, 2019.

\bibitem{Zeng2021RethinkingTB}
Yi~Zeng, Won Park, Zhuoqing~Morley Mao, and R.~Jia.
\newblock Rethinking the backdoor attacks’ triggers: A frequency perspective.
\newblock {\em 2021 IEEE/CVF International Conference on Computer Vision
  (ICCV)}, pages 16453--16461, 2021.

\bibitem{zhao2019seeing}
Yue Zhao, Hong Zhu, Ruigang Liang, Qintao Shen, Shengzhi Zhang, and Kai Chen.
\newblock Seeing isn't believing: Towards more robust adversarial attack
  against real world object detectors.
\newblock In {\em Proceedings of the 2019 ACM SIGSAC Conference on Computer and
  Communications Security}, pages 1989--2004, 2019.

\end{thebibliography}

\end{document}